\def\year{2019}\relax
\documentclass[letterpaper]{article} 
\usepackage{aaai19}  
\usepackage{times}  
\usepackage{helvet}  
\usepackage{courier}  
\usepackage{url}  
\usepackage{graphicx}  
\usepackage[colorlinks, citecolor=green]{hyperref}

\usepackage{mathtools}
\usepackage{amsthm}
\usepackage{subcaption}
\usepackage{paralist}
\usepackage{bm}
\usepackage{amsfonts}
\usepackage{multirow}
\usepackage{cases}
\usepackage{booktabs}
\usepackage{threeparttable}
\usepackage{epstopdf}
\usepackage{upgreek}
\usepackage{endnotes}
\usepackage{etoolbox}
\usepackage{algpseudocode}
\usepackage{amsmath}

\begin{document}
%
\title{Joint Domain Alignment and Discriminative Feature Learning for Unsupervised Deep Domain Adaptation\thanks{This work was supported by the opening foundation of the State Key Laboratory (No. 2014KF06), and the National Science and Technology Major Project (No. 2013ZX03005013).}}
\author{Chao Chen\thanks{Corresponding Author}, Zhihong Chen\thanks{Chao Chen and Zhihong Chen contributed equally to this work.}, Boyuan Jiang, Xinyu Jin\\
Institute of Information Science and Electronic Engineering\\
Zhejiang University, Hangzhou, China\\
\{chench,zhchen,byjiang,jinxy\}@zju.edu.cn
}
\maketitle

\begin{abstract}
Recently, considerable effort has been devoted to deep domain adaptation in computer vision and machine learning communities. However, most of existing work only concentrates on learning shared feature representation by minimizing the distribution discrepancy across different domains. Due to the fact that all the domain alignment approaches can only reduce, but not remove the domain shift, target domain samples distributed near the edge of the clusters, or far from their corresponding class centers are easily to be misclassified by the hyperplane learned from the source domain. To alleviate this issue, we propose to joint domain alignment and discriminative feature learning, which could benefit both domain alignment and final classification. Specifically, an instance-based discriminative feature learning method and a center-based discriminative feature learning method are proposed, both of which guarantee the domain invariant features with better intra-class compactness and inter-class separability. Extensive experiments show that learning the discriminative features in the shared feature space can significantly boost the performance of deep domain adaptation methods.
\end{abstract}

\section{Introduction}
Domain adaptation, which focuses on the issues of how to adapt the learned classifier from a source domain with a large amount of labeled samples to a target domain with limited or no labeled target samples even though the source and target domains have different, but related distributions, has received more and more attention in recent years. According to \cite{pan2010survey,csurka2017domain}, there are three commonly used domain adaptation approaches: feature-based domain adaptation, instance-based domain adaptation and classifier-based domain adaptation. The feature-based methods, which aim to learn a shared feature representation by minimizing the distribution discrepancy across different domains, can be further distinguished by: (a) the considered class of transformations \cite{gong2012geodesic,hoffman2013efficient,sun2016return}, (b) the types of discrepancy metrics, such as Maximum Mean Discrepancy (MMD) \cite{long2014transfer,tzeng2014deep,long2017deep}, Correlation Alignment (CORAL) \cite{sun2016return,sun2016deep}, Center Moment Discrepancy (CMD) \cite{zellinger2017central}, etc. The instance reweighting (also called landmarks selection) is another typical strategy for domain adaptation \cite{chu2013selective,hubert2016learning}, which considers that some source instances may not be relevant to the target even in the shared subspace. Therefore, it minimizes the distribution divergence by reweighting the source samples or selecting the landmarks, and then learns from those samples that are more similar to the target samples. Apart of this, the classifier-based domain adaptation represents another independent line of work \cite{yang2007adapting,rozantsev2018beyond,rozantsev2018residual,chen2018parameter}, which adapts the source model to the target by regularizing the difference between the source and target model parameters.
\begin{figure}
   \centering
     \includegraphics*[width=1.0\linewidth]{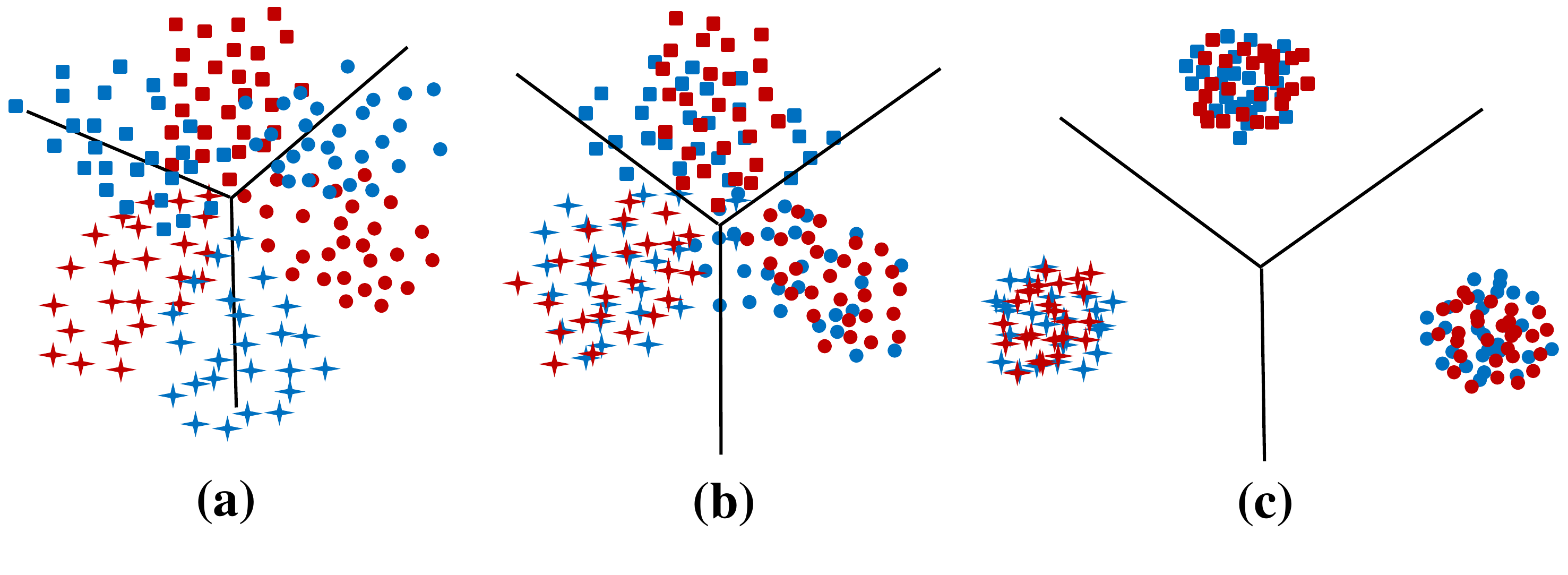}
   \caption{The necessity of joint domain alignment and discriminant features learning. Red: Source samples; Blue: Target samples; Black line: Hyperplane learned from the source domain; Circle, Square and Star indicate three different categories, respectively. (a) Source Only, due to the domain shift, the hyperplane learned from the source samples will misclassify a large amount of target samples. (b) Domain Alignment Only, the domain shift has been greatly reduced, but not removed, by the domain alignment. Therefore, the hyperplane learned from the source will still misclassify a few target samples which are almost distributed near the edge of the clusters, or far from their corresponding class centers. (c) Joint Domain Alignment and Discriminative Feature Learning, the hyperplane learned from the source can perfectly classify the target samples due to the discriminative-ness of the domain invariant features. (Best Viewed in Color)}
   \label{Fig1}
\end{figure}
\\\indent As mentioned above, the most recent work only devote to mitigating the domain shift by domain alignment. However, all the domain alignment approaches can only reduce, but not remove, the domain discrepancy. Therefore, the target samples distributed near the edge of the clusters, or far from their corresponding class centers are most likely to be misclassified by the hyperplane learned from the source domain. To alleviate this issue, a practical way is to enforce the target samples with better intra-class compactness. In this way, the number of samples that are far from the high density region and easily to be misclassified will be greatly reduced. Similarly, another feasible measure is to eliminate the harmful effects of the domain mismatch in the aligned feature space by enlarging the difference across different categories. However, under the unsupervised domain adaptation setting, it is quite difficult and inaccurate to obtain the category or cluster information of the target samples. Therefore, to enforce the target samples with better intra-class compactness and inter-class separability directly is somehow a hard work. Fortunately, recall that the source domain and target domain are highly-related and have similar distributions in the shared feature space. In this respect, it is reasonable to make the source features in the aligned feature space more discriminative, such that the target features maximally aligned with the source domain will become discriminative automatically.
\\\indent In this work, we propose to \textbf{j}oint domain alignment and \textbf{d}iscriminative feature learning for unsupervised deep \textbf{d}omain \textbf{a}daptation (\textbf{JDDA}). As can be seen in Fig. \ref{Fig1}, we  illustrate the necessity of joint domain alignment and discriminative feature learning. The merits of this paper include:\\
(1) As far as we know, this is the first attempt to jointly learn the discriminative deep features for deep domain adaptation.\\
(2) Instance-based and center-based discriminative learning strategies are proposed to learn the deep discriminative features.\\
(3) We analytically and experimentally demonstrate that the incorporation of the discriminative shared representation will further mitigate the domain shift and benefit the final classification, which would significantly enhance the transfer performance.

\section{Related Work}
Recently, a great deal of efforts have been made for domain adaptation based on the deep architectures. Among them, most of the deep domain adaptation methods follow the Siamese CNN architectures with two streams, representing the source model and target model respectively. In \cite{tzeng2014deep,long2015learning,sun2016deep}, the two-stream CNN shares the same weights between the source and target models, while \cite{rozantsev2018beyond,rozantsev2018residual} explores the two-stream CNN with related but non-shared parameters. As concluded in \cite{csurka2017domain}, the most commonly used deep domain adaptation approaches can be roughly classified into three categories: (1) Discrepancy-based methods, (2) Reconstruction-based methods and (3) Adversarial adaptation methods.

\indent The typical discrepancy-based methods can be seen in \cite{tzeng2014deep,long2015learning,sun2016deep}. They are usually achieved by adding an additional loss to minimize the distribution discrepancy between the source and target domains in the shared feature space. Specially, Zeng et al. \cite{tzeng2014deep} explores the Maximum Mean Discrepancy (MMD) to align the source and target domains, while Long et al. extend the MMD to multi-kernel MMD \cite{long2015learning,long2017deep} which aligns the joint distributions of multiple domain-specific layers across domains. Another impressive work is DeepCORAL \cite{sun2016deep}, which extends the CORAL to deep architectures, and aligns the covariance of the source and target features. Besides, the recently proposed Center Moment Discrepancy (CMD) \cite{zellinger2017central} diminishes the domain shift by aligning the central moment of each order across domains.

\indent Another important line of work is the reconstruction-based deep domain adaptation \cite{ghifary2016deep}, which jointly learns the shared encoding representation for both source label prediction and unlabeled target samples reconstruction. In contrast, domain separation networks (DSN) \cite{bousmalis2016domain} introduce the notion of a private subspace for each domain, which captures domain specific properties using the encoder-decoder architectures. Besides, Tan et al. propose a Selective Learning Algorithm (SLA) \cite{tan2017distant}, which gradually selects the useful unlabeled data from the intermediate domains using the reconstruction error. The adversarial adaptation method is another increasingly popular approach. The representative work is to optimize the source and target mappings using the standard minimax objective \cite{ganin2015unsupervised,ganin2016domain}, the symmetric confusion objective \cite{tzeng2015simultaneous} or the inverted label objective \cite{tzeng2017adversarial}.

\indent Recently, there is a trend to improve the performance of CNN by learning even more discriminative features. Such as contrastive loss \cite{sun2014deep} and center loss \cite{wen2016discriminative}, which are proposed to learn discriminative deep features for face recognition and face verification. Besides, the large-margin softmax loss (L-Softmax) \cite{liu2016large} is proposed to generalize the softmax loss to large margin softmax, leading to larger angular separability between learned features. Inspired by these methods, we propose two discriminative feature learning methods, i.e., Instance-Based discriminative feature learning and Center-Based discriminative feature learning. By jointing domain alignment and discriminative feature learning, the shared representations could be better clustered and more separable, which can evidently contribute to domain adaptation.

\section{Our Approach}
\begin{figure}
   \centering
     \includegraphics*[width=0.55\linewidth,angle=90]{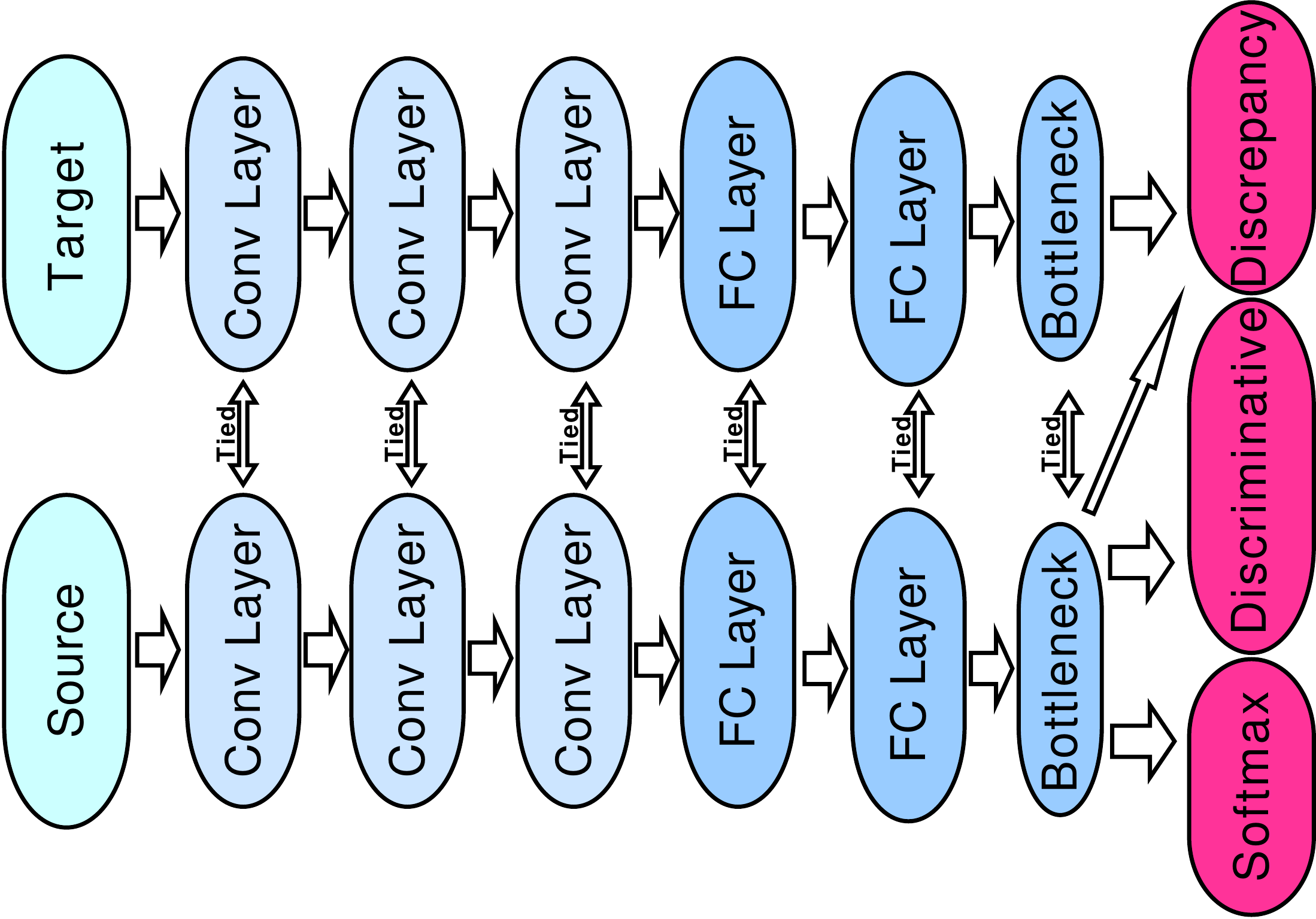}
   \caption{The proposed two-stream CNN for domain adaptation. We introduce a discriminative loss, which enforces the domain invariant features with smaller intra-class scatter and better inter-class separability. Note that both the domain discrepancy loss and the discriminative loss are applied in the bottleneck layer.}
   \label{Fig2}
\end{figure}

In this section, we present our proposed \textbf{JDDA} in detail. Following their work \cite{tzeng2014deep,long2017deep,sun2016deep}, the two-stream CNN architecture with shared weights is adopted. As illustrated in Fig. \ref{Fig2}, the first stream operates the source data and the second stream operates the target data. What distinguishes our work from others is that an extra discriminative loss is proposed to encourage the shared representations to be more discriminative, which is demonstrated to be good for both domain alignment and final classification.

\indent In this work, following the settings of unsupervised domain adaptation, we define the labeled source data as $\mathcal{D}^s=\{\mathbf{X}^s,\mathbf{Y}^s\}=\{(\bm{x}_i^s,y_i^s)\}_{i=1}^{n_s}$  and define the unlabeled target data as $\mathcal{D}^t=\{\mathbf{X}^t\}=\{\bm{x}_i^t\}_{i=1}^{n_t}$, where $\mathbf{x}^s$ and $\mathbf{x}^t$ have the same dimension $\mathbf{x}^{s(t)}\in\mathbb{R}^d$. Let $\bm\Theta$ denotes the shared parameters to be learned. $\mathbf{H}_s\in\mathbb{R}^{b\times L}$ and $\mathbf{H}_t\in\mathbb{R}^{b\times L}$ denote the learned deep features in the bottleneck layer regard to the source stream and target stream, respectively. $b$ indicates the batch size during the training stage and $L$ is the number of hidden neurons in the bottleneck layer. Then, the networks can be trained by minimizing the following loss function.
\begin{equation}\label{Eq1}
\mathcal{L}(\bm\Theta|\mathbf{X}_s,\mathbf{Y}_s,\mathbf{X}_t)=\mathcal{L}_s+\lambda_1\mathcal{L}_c+\lambda_2\mathcal{L}_d
\end{equation}
\begin{equation}\label{Eq2}
\mathcal{L}_s=\dfrac{1}{n_s}\sum_{i=1}^{n_s}c(\bm\Theta|\bm{x}_i^s,y_i^s)
\end{equation}
\begin{equation}\label{Eq3}
\mathcal{L}_c=CORAL(\mathbf{H}_s,\mathbf{H}_t)
\end{equation}
\begin{equation}\label{Eq4}
\mathcal{L}_d=\mathcal{J}_d(\bm\Theta|\mathbf{X}^s,\mathbf{Y}^s)
\end{equation}
Here, $\mathcal{L}_s$, $\mathcal{L}_c$ and $\mathcal{L}_d$ represent the source loss, domain discrepancy loss and discriminative loss, respectively. $\lambda_1$ and $\lambda_2$ are trade-off parameters to balance the contributions of the domain discrepancy loss and the discriminative loss. Specifically, $c(\bm\Theta|\bm{x}_i^s,y_i^s)$ denotes the standard classification loss with respect to the source data. $\mathcal{L}_c=CORAL(\mathbf{H}_s,\mathbf{H}_t)$ denotes the domain discrepancy loss measured by the correlation alignment (CORAL) \cite{sun2016return,sun2016deep}. $\mathcal{J}_d(\bm\Theta|\mathbf{X}^s,\mathbf{Y}^s)$ indicates our proposed discriminative loss, which guarantees the domain invariant features with better intra-class compactness and inter-class separability.

\subsection{Correlation Alignment}
To learn the domain invariant features, the CORAL is adopted, which diminishes the domain discrepancy by aligning the covariance of the source and target features. The domain discrepancy loss measured by CORAL can be expressed as
\begin{equation}\label{Eq5}\small
\mathcal{L}_c=CORAL(\mathbf{H}_s,\mathbf{H}_t)=\dfrac{1}{4L^2}\Vert Cov(\mathbf{H}_s)-Cov(\mathbf{H}_t)\Vert_F^2
\end{equation}
where $\Vert\bm\cdot\Vert_F^2$ denotes the squared matrix Frobenius norm. $Cov(\mathbf{H}_s)$ and $Cov(\mathbf{H}_t)$ denote the covariance matrices of the source and target features in the bottleneck layer, which can be computed as
$Cov(\mathbf{H}_s)=\mathbf{H}_s^{\top}\mathbf{J}_b\mathbf{H}_s$, and $Cov(\mathbf{H}_t)=\mathbf{H}_t^{\top}\mathbf{J}_b\mathbf{H}_t$. $\mathbf{J}_b=\mathbf{I}_b-\tfrac{1}{b}\mathbf{1}_n\mathbf{1}_n^{\top}$ is the centralized matrix, where $\mathbf{1}_b\in\mathbb{R}^b$ is an all-one column vector, and $b$ is the batch-size. Note that the training process is implemented by mini-batch SGD, therefore, only a batch of training samples are aligned in each iteration. Interested readers may refer \cite{sun2016return,sun2016deep} for more details.

\subsection{Discriminative Feature Learning}
In order to enforce the two-stream CNN to learn even more discriminative deep features, we propose two discriminative feature learning methods, i.e., the Instance-Based discriminative feature learning and the Center-Based discriminative feature learning. It is worth noting that the whole training stage is based on the mini-batch SGD. Therefore, the discriminative loss presented below is also based on a batch of samples.

\subsubsection{Instance-Based Discriminative Loss}
The motivation of the Instance-Based discriminative feature learning is that the samples from the same class should be as closer as possible in the feature space, and the samples from different classes should be distant from each other by a large margin. In this respect, the Instance-Based discriminative loss $\mathcal{L}_d^{I}$ can be formulated as
\begin{equation}\label{Eq6}\small
\mathcal{J}_d^{I}(\mathbf{h}_i^s,\mathbf{h}_j^s)=
\begin{cases}
\max(0,\Vert\mathbf{h}_i^s-\mathbf{h}_j^s\Vert_2-m_1)^2   & C_{ij}=1 \\
\max(0,m_2-\Vert\mathbf{h}_i^s-\mathbf{h}_j^s\Vert_2)^2  & C_{ij}=0
\end{cases}
\end{equation}
\begin{equation}\label{Eq6_1}
\mathcal{L}_d^{I}=\sum_{i,j=1}^{n_s}\mathcal{J}_d^{I}(\mathbf{h}_i^s,\mathbf{h}_j^s)
\end{equation}
where $\mathbf{h}_i^s\in\mathbb{R}^L$ ($L$ is the number of neurons in the bottleneck layer) denotes the $i$-th deep feature of bottleneck layer w.r.t. the $i$-th training sample, and $\mathbf{H}_s=[\mathbf{h}_1^s;\mathbf{h}_2^s;\cdots;\mathbf{h}_b^s]$. $C_{ij}=1$ means that $\mathbf{h}_i^s$ and $\mathbf{h}_j^s$ are from the same class, and $C_{ij}=0$ means that $\mathbf{h}_i^s$ and $\mathbf{h}_j^s$ are from different classes. As can be seen in \eqref{Eq6}\eqref{Eq6_1}, the discriminative loss will enforce the distance between intra-class samples no more than $m_1$ and the distance between the paired inter-class samples at least $m_2$. Intuitively, this penalty will undoubtedly enforce the deep features to be more discriminative. For brevity, we denote the pairwise distance of the deep features $\mathbf{H}_s$ as $\mathbf{D}^H\in\mathbb{R}^{b\times b}$, where $\mathbf{D}^H_{ij}=\Vert\mathbf{h}_i^s-\mathbf{h}_j^s\Vert_2$. Let $\mathbf{L}\in\mathbb{R}^{b\times b}$ denotes the indictor matrix, $\mathbf{L}_{ij}=1$ if the $i$-th and $j$-th samples are from the same class and $\mathbf{L}_{ij}=0$ if they are from different classes. Then, the Instance-Based discriminative loss can be simplified to
\begin{equation}\label{Eq7}
\begin{split}
\mathcal{L}_d^I&=\upalpha\Vert\max(0,\mathbf{D}^H-m_1)^2\circ\mathbf{L}\Vert_{sum} \\
&+\Vert\max(0,m_2-\mathbf{D}^H)^2\circ(1-\mathbf{L})\Vert_{sum}
\end{split}
\end{equation}
where the square operate denotes element-wise square and "$\circ$" denotes element-wise multiplication. $\Vert\bm\cdot\Vert_{sum}$ represents the sum of all the elements in the matrix. $\alpha$ is the trade-off parameter introduced to balance the intra-class compactness and inter-class separability. Note that the Instance-Based discriminative learning method is quite similar with the manifold embedding \cite{weston2012deep} related methods. Both of them encourage the similar samples to be closer and dissimilar samples to be far from each other in the embedding space. The difference is that the similarity in our proposal is defined by the labels, while the manifold embedding is an unsupervised approach and defines the similarity by the distance in the input space.

\subsubsection{Center-Based Discriminative Loss} To calculate the Instance-Based discriminative loss, the calculation of pairwise distance is required, which is computationally expensive. Inspired by the Center Loss \cite{wen2016discriminative} which penalizes the distance of each sample to its corresponding class center, we proposed the Center-Based discriminative feature learning as below.
\begin{equation}\label{Eq8}
\begin{split}
\mathcal{L}_d^C&=\upbeta\sum_{i=1}^{n_s}\max(0,\Vert\mathbf{h}_i^s-\mathbf{c}_{y_i}\Vert_2^2-m_1)+ \\
&\sum_{i,j=1,i\neq j}^c\max(0,m_2-\Vert\mathbf{c}_i-\mathbf{c}_j\Vert_2^2)
\end{split}
\end{equation}
where $\upbeta$ is the trade-off parameter, $m_1$ and $m_2$ are two constraint margins. The $\mathbf{c}_{y_i}\in\mathbb{R}^d$ denotes the $y_i$-th class center of the deep features, $y_i\in\{1,2,\cdots,c\}$ and $c$ is the number of class. Ideally, the class center $\mathbf{c}_i$ should be calculated by averaging the deep features of all the samples. Due to the fact that we perform the update based on mini-batch, it is quite difficult to average the deep features by the whole training set. Herein, we make a necessary modification. For the second term of the discriminative loss in \eqref{Eq8}, the $\mathbf{c}_i$ and $\mathbf{c}_j$ used to measure the inter-class separability are approximately computed by averaging the current batch of deep features, which we call the "batch class center". Instead, the $\mathbf{c}_{y_i}$ used to measure the intra-class compactness should be more accurate and closer to the "global class center". Therefore, we updated the $\mathbf{c}_{y_i}$ in each iteration as
\begin{equation}\label{Eq9}
\Delta\mathbf{c}_j=\dfrac{\sum_{i=1}^b\delta(y_i=j)(\mathbf{c}_j-\mathbf{h}_i^s)}{1+\sum_{i=1}^b\delta(y_i=j)}
\end{equation}
\begin{equation}\label{Eq10}
\mathbf{c}_j^{t+1}=\mathbf{c}_j^t-\gamma\cdot\Delta\mathbf{c}_j^t
\end{equation}
The "global class center" is initialized as the "batch class center" in the first iteration and updated according to the coming batch of samples via \eqref{Eq9}\eqref{Eq10} in each iteration, where $\gamma$ is the learning rate to update the "global class center". For brevity, \eqref{Eq8} can be simplified to
\begin{equation}\label{Eq11}
\begin{split}
\mathcal{L}_d^C=\upbeta\Vert\max(0,\mathbf{H}^c-m_1)\Vert_{sum}+ \\
\Vert\max(0,m_2-\mathbf{D}^c)\circ\mathbf{M}\Vert_{sum}
\end{split}
\end{equation}
where $\mathbf{H}^c=[\mathbf{h}_1^c;\mathbf{h}_2^c;\dots;\mathbf{h}_b^c]$ has the same size as $\mathbf{H}_s$, and $\mathbf{h}_i^c=\Vert\mathbf{h}_i^s-\mathbf{c}_{y_i}\Vert_2^2$ denotes the distance between the $i$-th deep feature $\mathbf{h}_i^s$ and its corresponding center $\mathbf{c}_{y_i}$. $\mathbf{D}^c\in\mathbb{R}^{c\times c}$ denotes the pairwise distance of the "batch class centers", i.e., $\mathbf{D}_{ij}^c=\Vert\mathbf{c}_i-\mathbf{c}_j\Vert_2^2$. $\mathbf{M}=\mathbf{1}_b\mathbf{1}_b^{\top}-\mathbf{I}_b$, and "$\circ$" denotes the element-wise multiplication. Different from the Center Loss, which only considers the intra-class compactness. Our proposal not only  penalizes the distances between the deep features and their corresponding class centers, but also enforces large margins among centers across different categories.

\subsubsection{Discussion}
Whether it is Instance-Based method or Center-Based method, it can make the deep features more discriminative. Besides, these two methods can be easily implemented and integrated into modern deep learning frameworks. Compared with the Instance-Based method, the computation of the Center-Based method is more efficient. Specifically, The  computational complexity of  Center-Based method is theoretically $O(n_sc+c^2)$ and $O(bc+c^2)$ when using mini-batch SGD, while the Instance-Based method needs to compute the pairwise distance, therefore, its complexity is $O(n_s^2)$ in theory and $O(b^2)$ when using mini-batch SGD. Besides, the Center-Based method should converge faster intuitively (This can also be evidenced in our experiments), because it takes the global information into consideration in each iteration, while the Instance-Based method only regularizes the distance of pairs of instances.

\subsection{Training}
Both the proposed Instance-Based joint discriminative domain adaptation (\textbf{JDDA-I}) and Center-Based joint discriminative domain adaptation (\textbf{JDDA-C}) can be easily implemented via the mini-batch SGD. For the \textbf{JDDA-I}, the total loss is given as $\mathcal{L}=\mathcal{L}_s+\lambda_1\mathcal{L}_c+\lambda_2^I\mathcal{L}_d^I$, while the source loss is defined by the conventional softmax classifier. $\mathcal{L}_c$ defined in \eqref{Eq5} and $\mathcal{L}_d^I$ defined in \eqref{Eq7} are both differentiable w.r.t. the inputs. Therefore, the parameters $\bm\Theta$ can be directly updated by the standard back propagation
\begin{equation}
\bm\Theta^{t+1}=\bm\Theta^t-\eta\dfrac{\partial(\mathcal{L}_s+\lambda_1\mathcal{L}_c+\lambda_2^I\mathcal{L}_d^I)}{\partial\mathbf{x_i}}
\end{equation}
where $\eta$ is the learning rate. Since the "global class center" can not be computed by a batch of samples, the \textbf{JDDA-C} has to update $\bm\Theta$ as well as the "global class center" simultaneously in each iteration. i.e.,
 \begin{equation}
\bm\Theta^{t+1}=\bm\Theta^t-\eta\dfrac{\partial(\mathcal{L}_s+\lambda_1\mathcal{L}_c+\lambda_2^C\mathcal{L}_d^C)}{\partial\mathbf{x_i}}
\end{equation}
\begin{equation}\label{Eq14}
\mathbf{c}_j^{t+1}=\mathbf{c}_j^t-\gamma\cdot\Delta\mathbf{c}_j^t  \quad j=1,2,\cdots,c
\end{equation}

\section{Experiments}
In this section, we evaluate the efficacy of our approach by comparing against several state-of-the-art deep domain adaptation methods on two image classification adaptation datasets, one is the Office-31 dataset \cite{Saenko2010Adapting}, the other is a large-scale digital recognition dataset. The source code of the JDDA is released online\footnote{\url{https://github.com/chenchao666/JDDA-Master}}

\subsection{Setup}
\subsubsection{Office-31}
is a standard benchmark for domain adaptation in computer vision, comprising 4,110 images in 31 classes collected from three distinct domains: Amazon (A), which contains images downloaded from amazon.com, Webcam (W) and DSLR (D), which contain images taken by web camera and digital SLR camera with different photographical settings, respectively. We evaluate all methods across all six transfer tasks \textbf{A}$\rightarrow$\textbf{W}, \textbf{W}$\rightarrow$\textbf{A}, \textbf{W}$\rightarrow$\textbf{D}, \textbf{D}$\rightarrow$\textbf{W}, \textbf{A}$\rightarrow$\textbf{D} and \textbf{D}$\rightarrow$\textbf{A} as in \cite{long2015learning}. These tasks represent the performance on the setting where both source and target domains have small number of samples.

\subsubsection{Digital recognition dataset}
contains five widely used benchmarks: Street View House Numbers (SVHN) \cite{Netzer2011Reading}, MNIST \cite{Lecun1998Gradient}, MNIST-M \cite{ganin2016domain}, USPS \cite{Hull2002A} and synthetic digits dataset (syn digits) \cite{ganin2016domain}, which consist 10 classes of digits. We evaluate
 our approach over four cross-domain pairs:  \textbf{SVHN}$\rightarrow$ \textbf{MNIST}, \textbf{MNIST}$\rightarrow$\textbf{MNIST-M}, \textbf{MNIST}$\rightarrow$\textbf{USPS} and \textbf{synthetic digits}$\rightarrow$\textbf{MNIST}. Different from Office-31 where different domains are of small but different sizes, each of the five domains has a large-scale and a nearly equal number of samples, which makes it a good complement to Office-31 for more controlled experiments.

\subsubsection{Compared Approaches}
We mainly compare our proposal with Deep Domain Confusion (\textbf{DDC}) \cite{tzeng2014deep}, Deep Adaptation Network (\textbf{DAN}) \cite{long2015learning}, Domain Adversarial Neural Network (\textbf{DANN}) \cite{ganin2016domain}, Center Moment Discrepancy (\textbf{CMD}) \cite{zellinger2017central}, Adversarial Discriminative Domain Adaptation (\textbf{ADDA}) \cite{tzeng2017adversarial} and Deep Correlation Alignment (\textbf{CORAL}) \cite{sun2016return} since these approaches and our \textbf{JDDA} are all proposed for learning domain invariant feature representations.

\subsection{Implementation Details}
For the experiments on Office-31, both the compared approaches and our proposal are trained by fine-tuning the ResNet pre-trained on ImageNet, and the activations of the last layer \emph{pool5} are used as image representation. We follow standard evaluation for unsupervised adaptation \cite{long2015learning} and use all source examples with labels and all target examples without labels. For the experiments on digital recognition dataset, we use the modified LeNet to verify the effectiveness of our approach. We resize all images to 32 $\times$ 32 and convert RGB images to grayscale. For all transfer tasks, we perform five random experiments and report the averaged results across them. For fair comparison, all deep learning based models above have the same architecture as our approach for the label predictor.

Note that all the above methods are implemented via tensorflow and trained with Adam optimizer. When fine-tune the ResNet (50 layers), we only update the weights of the full-connected layers (\emph{fc}) and the final block (scale5/block3) and fix other layers due to the small sample size of the Office-31.
For each approach we use a batch size of 256 samples in total with 128 samples from each domain, and set the learning rate $\eta$ to $10^{-4}$ and the learning rate of "global class center" $\gamma$ to 0.5. When implementing the methods proposed by others, instead of fixing the adaptation factor $\lambda$, we gradually update it from 0 to 1 by a progressive schedule: $\lambda_p=\frac{2}{1+exp(-\mu p)}-1$, where $p$ is the training progress linearly changing from 0 to 1 and $\mu=10$ is fixed throughout experiments \cite{long2017deep}.  This progressive strategy reduces parameter sensitivity and eases the selection of models. As our approach can work stably across different transfer tasks, the hyper-parameter $\lambda_2$ is first selected according to accuracy on SVHN$\rightarrow$MNIST (results are shown in the Figure \ref{sensi2}) and then fixed as $\lambda_2^I$ = 0.03 (\textbf{JDDA-I}) and $\lambda_2^C$ = 0.01 (\textbf{JDDA-C}) for all other transfer tasks. We also fixed the constraint margins as $m_1 = 0$ and $m_2 = 100$ throughout experiments.

\begin{table*}[ht]
\centering
\caption{results (accuracy \%) on Office-31 dataset  for unsupervised domain adaptation based on ResNet}\label{tab:aStrangeTable}
\label{tab1}

\begin{tabular}{cccccccc}
\toprule
Method& A$\rightarrow$W&D$\rightarrow$W&W$\rightarrow$D&A$\rightarrow$D&D$\rightarrow$A&W$\rightarrow$A&Avg\\
\midrule
ResNet \cite{He_2016_CVPR}&73.1$\pm$0.2&93.2$\pm$0.2&$98.8\pm$0.1&72.6$\pm$0.2&55.8$\pm$0.1&56.4$\pm$0.3&75.0\\
DDC \cite{tzeng2014deep}&74.4$\pm$0.3&94.0$\pm$0.1&98.2$\pm$0.1&74.6$\pm$0.4&56.4$\pm$0.1&56.9$\pm$0.1&75.8\\
DAN \cite{long2015learning}&78.3$\pm$0.3&$\textbf{95.2}\pm$0.2&$99.0\pm$0.1&75.2$\pm$0.2&\textbf{58.9}$\pm$0.2&64.2$\pm$0.3&78.5\\
DANN \cite{ganin2016domain}&73.6$\pm$0.3&94.5$\pm$0.1&99.5$\pm$0.1&74.4$\pm$0.5&57.2$\pm$0.1&60.8$\pm$0.2&76.7\\
CMD \cite{zellinger2017central}&76.9$\pm$0.4&94.6$\pm$0.3&99.2$\pm$0.2&75.4$\pm$0.4&56.8$\pm$0.1&61.9$\pm$0.2&77.5\\
CORAL \cite{sun2016deep}&79.3$\pm$0.3&94.3$\pm$0.2&99.4$\pm$0.2&74.8$\pm$0.1&56.4$\pm$0.2&63.4$\pm$0.2&78.0\\
\textbf{JDDA-I} &82.1$\pm$0.3&\textbf{95.2}$\pm$0.1&\textbf{99.7}$\pm$0.0&76.1$\pm$0.2&56.9$\pm$0.0&65.1$\pm$0.3&79.2\\
\textbf{JDDA-C} &\textbf{82.6}$\pm$0.4&\textbf{95.2}$\pm$0.2&\textbf{99.7}$\pm$0.0&\textbf{79.8}$\pm$0.1&57.4$\pm$0.0&\textbf{66.7}$\pm$0.2&\textbf{80.2}\\
\bottomrule
\end{tabular}
\end{table*}

\begin{table*}[ht]
\centering
\caption{results (accuracy \%) on digital recognition dataset  for unsupervised domain adaptation based on modified LeNet}\label{tab:aStrangeTable}
\label{tab2}
\begin{tabular}{cccccc}
\toprule
Method& SVHN$\rightarrow$MNIST&MNIST$\rightarrow$MNIST-M&USPS$\rightarrow$MNIST&SYN$\rightarrow$MNIST&Avg\\
\midrule
Modified LeNet &67.3$\pm$0.3&62.8$\pm$0.2&$66.4\pm$0.4&89.7$\pm$0.2&71.6\\
DDC \cite{tzeng2014deep}&71.9$\pm$0.4&78.4$\pm$0.1&75.8$\pm$0.3&89.9$\pm$0.2&79.0\\
DAN \cite{long2015learning}&79.5$\pm$0.3&$79.6\pm$0.2&$89.8\pm$0.2&75.2$\pm$0.1&81.0\\
DANN \cite{ganin2016domain}&70.6$\pm$0.2&76.7$\pm$0.4&76.6$\pm$0.3&90.2$\pm$0.2&78.5\\
CMD \cite{zellinger2017central}&86.5$\pm$0.3&85.5$\pm$0.2&86.3$\pm$0.4&96.1$\pm$0.2&88.6\\
ADDA \cite{tzeng2017adversarial}&72.3$\pm$0.2&$80.7\pm$0.3&92.1$\pm$0.2&96.3$\pm$0.4&85.4\\
CORAL \cite{sun2016deep}&89.5$\pm$0.2&81.6$\pm$0.2&96.5$\pm$0.3&96.5$\pm$0.2&91.0\\
\textbf{JDDA-I} &93.1$\pm$0.2&87.5$\pm$0.3&\textbf{97.0}$\pm$0.2&97.4$\pm$0.1&93.8\\
\textbf{JDDA-C} &\textbf{94.2}$\pm$0.1&\textbf{88.4}$\pm$0.2&96.7$\pm$0.1&\textbf{97.7}$\pm$0.0&\textbf{94.3}\\
\bottomrule
\end{tabular}
\end{table*}
\subsection{Result and Discussion}
The unsupervised adaptation results on the Office-31 dataset based on ResNet are shown in Table \ref{tab1}. As can be seen, our proposed JDDA outperforms all comparison methods on most transfer tasks. It is worth nothing that our approach improves the classification accuracy substantially on hard transfer tasks, e.g. \textbf{A}$\rightarrow$\textbf{W} where the source and target domains are remarkably different and \textbf{W}$\rightarrow$\textbf{A} where the size of the source domain is even smaller than the target domain, and achieves comparable performance on the easy transfer tasks, \textbf{D}$\rightarrow$\textbf{W} and \textbf{W}$\rightarrow$\textbf{D}, where source and target domains are similar. Thus we can draw a conclusion that our proposal has the ability to learn more transferable representations and can be applied to small-scale datasets adaption effectively by using a pre-trained deep model.

The results reveal several interesting observations. \textbf{(1)} Discrepancy-based methods achieve better performance than standard deep learning method (ResNet), which confirms that embedding domain-adaption modules into deep networks (DDC, DAN, CMD, CORAL) can reduce the domain discrepancy and improve domain adaptation performance. \textbf{(2)} Adversarial adaptation methods (DANN, ADDA) outperform source-only method, which validates that adversarial training process can learn more transferable features and thus improve the generalization performance. \textbf{(3)} The JDDA model performs the best and sets new state-of-the-art results. Different from all previous deep transfer learning methods that
the distance relationship of the source samples in the feature space is unconsidered during training, we add a discriminative loss using the information of the source domain labels, which explicitly encourages intra-class compactness and inter-class separability among learned features.

In contrast to Office-31 dataset, digital recognition dataset has a much larger domain size. With these large-scale transfer tasks, we are expecting to testify whether domain adaptation improves when domain sizes is large. Table \ref{tab2} shows the classification accuracy results based on the modified LeNet. We observe that JDDA outperforms all comparison methods on all the tasks. In particular, our approach improves the accuracy by a huge margin on difficult transfer tasks, e.g. SVHN$\rightarrow$MNIST and MNIST$\rightarrow$MNIST-M. In the task of SVHN$\rightarrow$MNIST, the SVHN dataset contains significant variations (in scale, background clutter, blurring, slanting, contrast and rotation) and there is only slightly variation in the actual digits shapes, that makes it substantially different from MNIST. In the domain adaption scenario of MNIST$\rightarrow$MNIST-M. The MNIST-M quite distinct from the dataset of MNIST, since it was created by using each MNIST digit as a binary mask and inverting with it the colors of a background image randomly cropped from the Berkeley Segmentation Data Set (BSDS500) \cite{Arbel2011Contour}. The above results suggest that the proposed discriminative loss $\mathcal{L}_d$ is also effective for large-scale domain adaption.
\subsection{Analysis}
\begin{figure*}
    \centering
  \begin{subfigure}[b]{0.248\textwidth}
    \includegraphics[width=\textwidth]{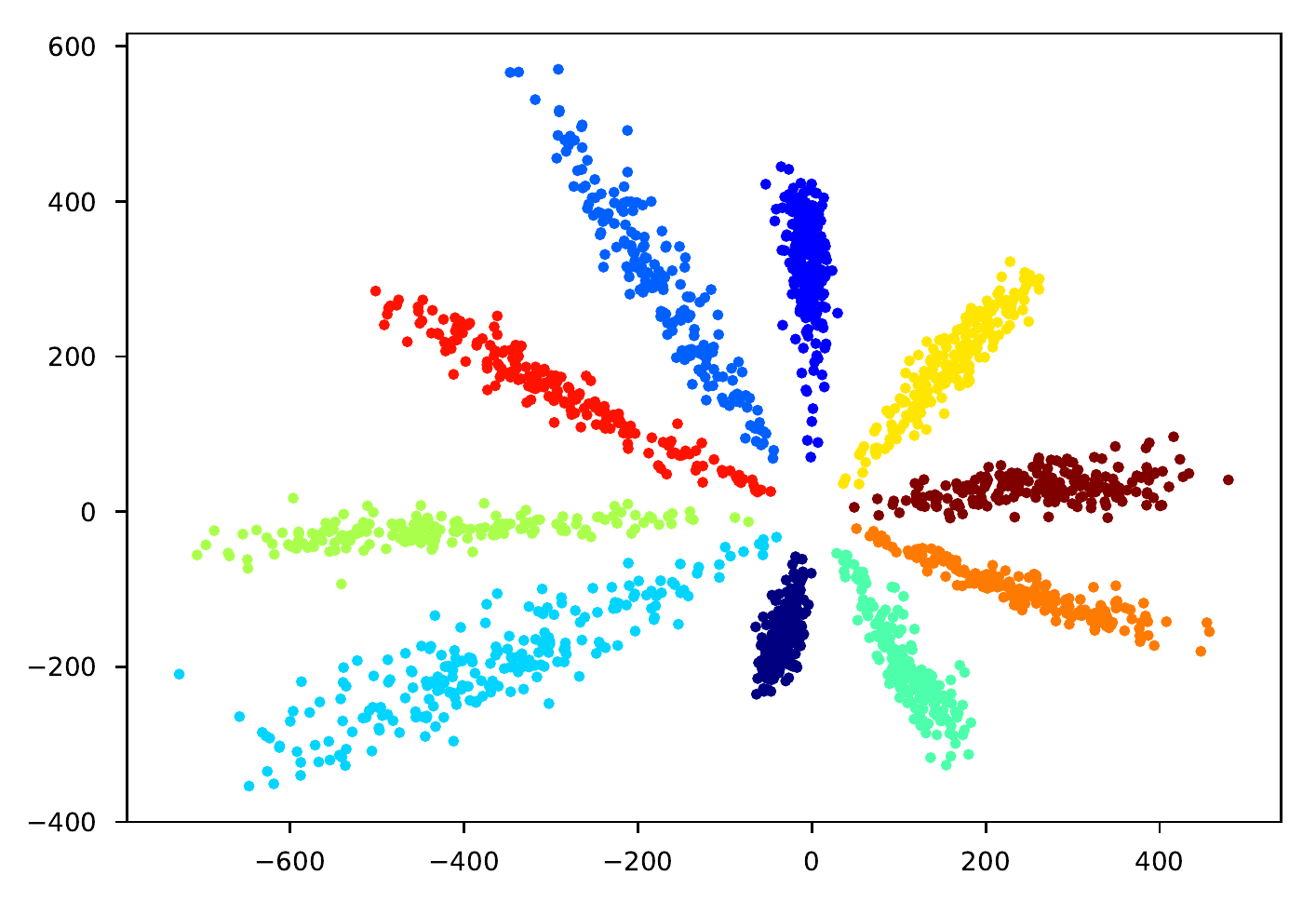}
    \caption{$\mathcal{L}_s$ (2D)}
    \label{2D1}
  \end{subfigure}
   \begin{subfigure}[b]{0.248\textwidth}
    \includegraphics[width=\textwidth]{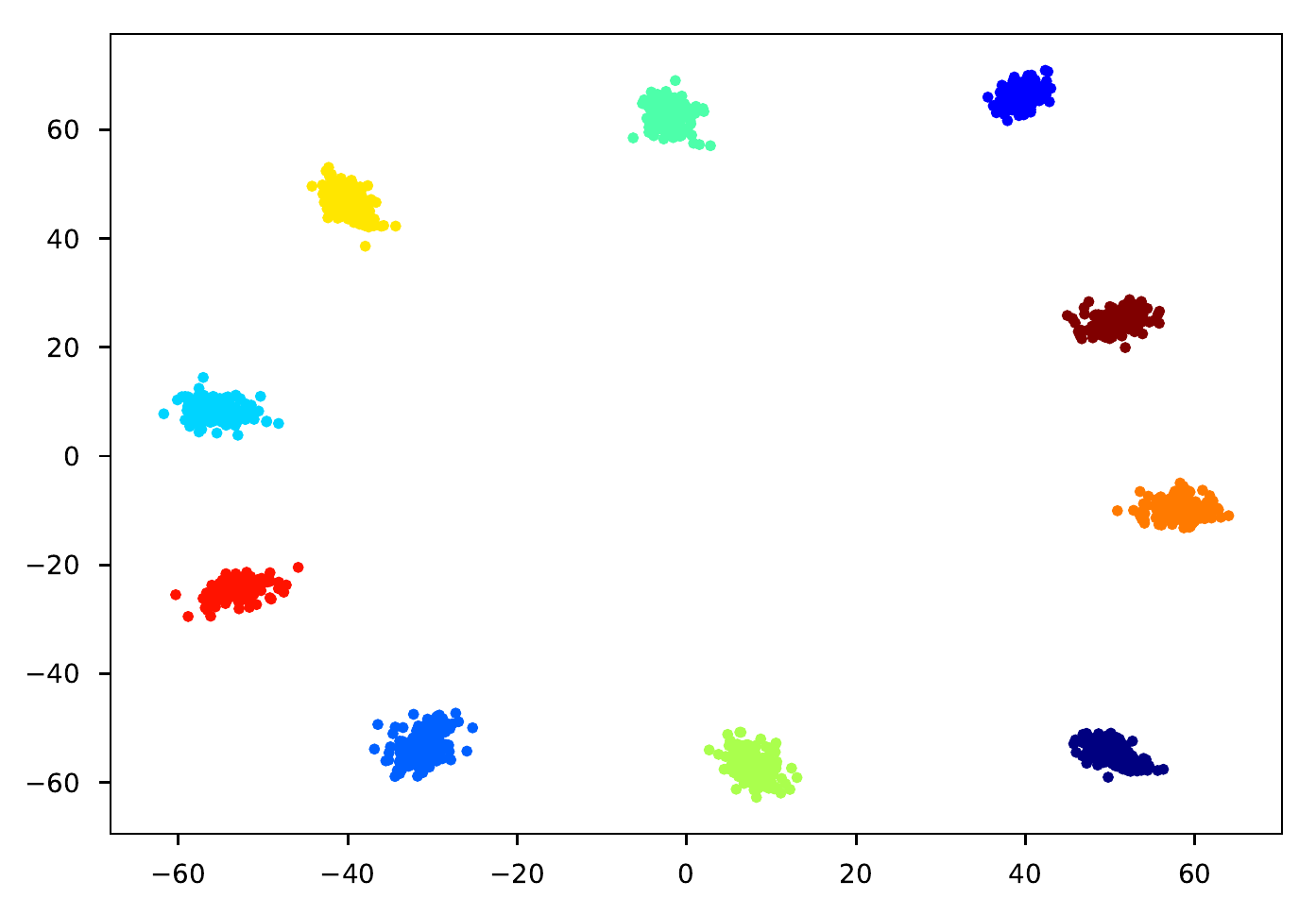}
    \caption{$\mathcal{L}_s$ + $\mathcal{L}_d$ (2D)}
    \label{2D2}
  \end{subfigure}
    \begin{subfigure}[b]{0.245\textwidth}
    \includegraphics[width=\textwidth]{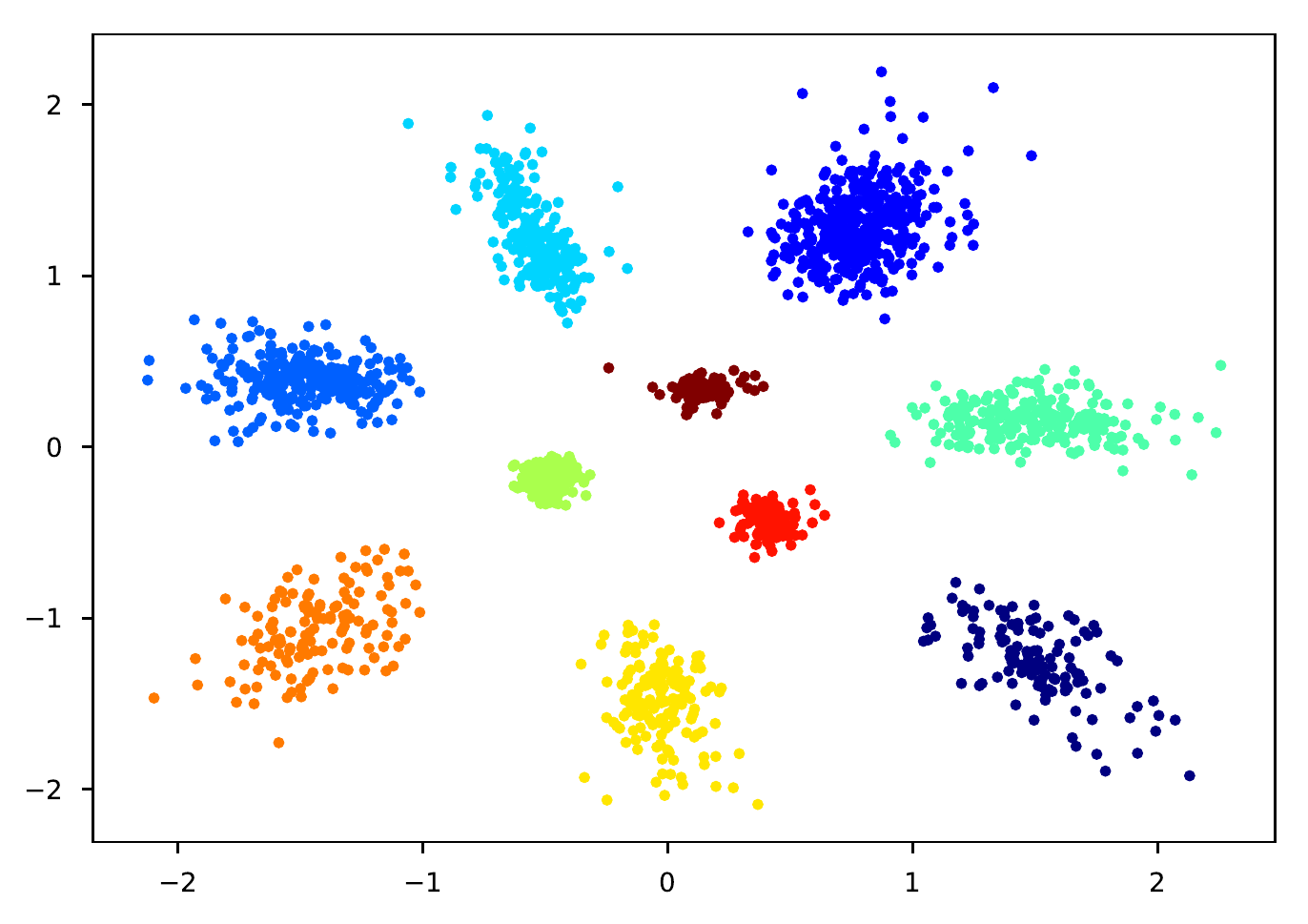}
    \caption{$\mathcal{L}_s$ + $\mathcal{L}_c$ (2D)}
    \label{2D3}
  \end{subfigure}
    \begin{subfigure}[b]{0.245\textwidth}
    \includegraphics[width=\textwidth]{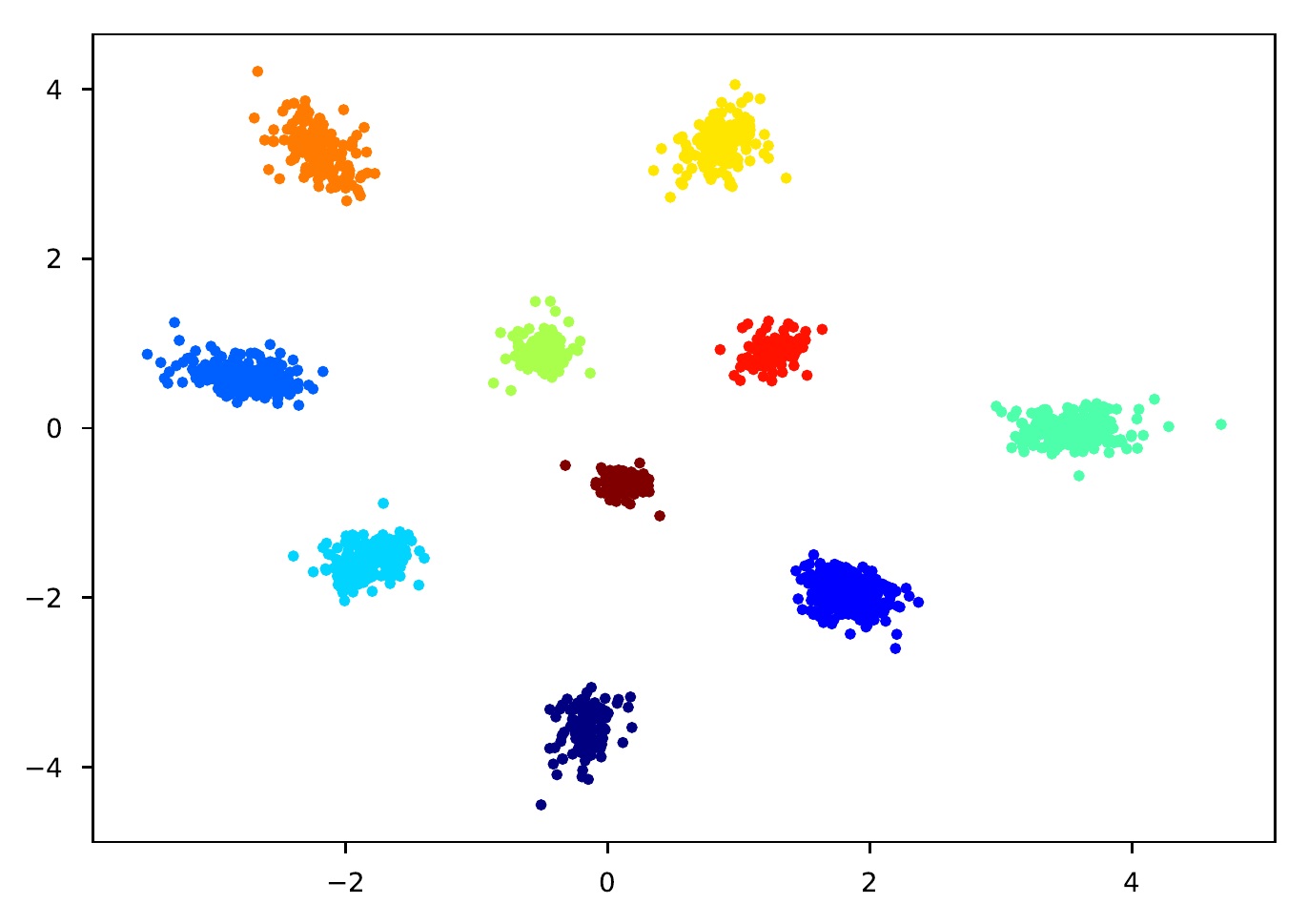}
    \caption{$\mathcal{L}_s$ + $\mathcal{L}_c$ + $\mathcal{L}_d$ (2D)}
    \label{2D4}
  \end{subfigure}
\caption{ features visualization (without our discriminative loss (a)(c) VS. with our discriminative loss (b)(d)) in SVHN dataset. It is worth noting that we set the feature (input of the Softmax loss) dimension as 2, and then plot them by class information. }
\label{img2DD}
\end{figure*}
\begin{figure*}[!ht]
\centering
  \begin{subfigure}[b]{0.245\textwidth}
    \includegraphics[width=\textwidth]{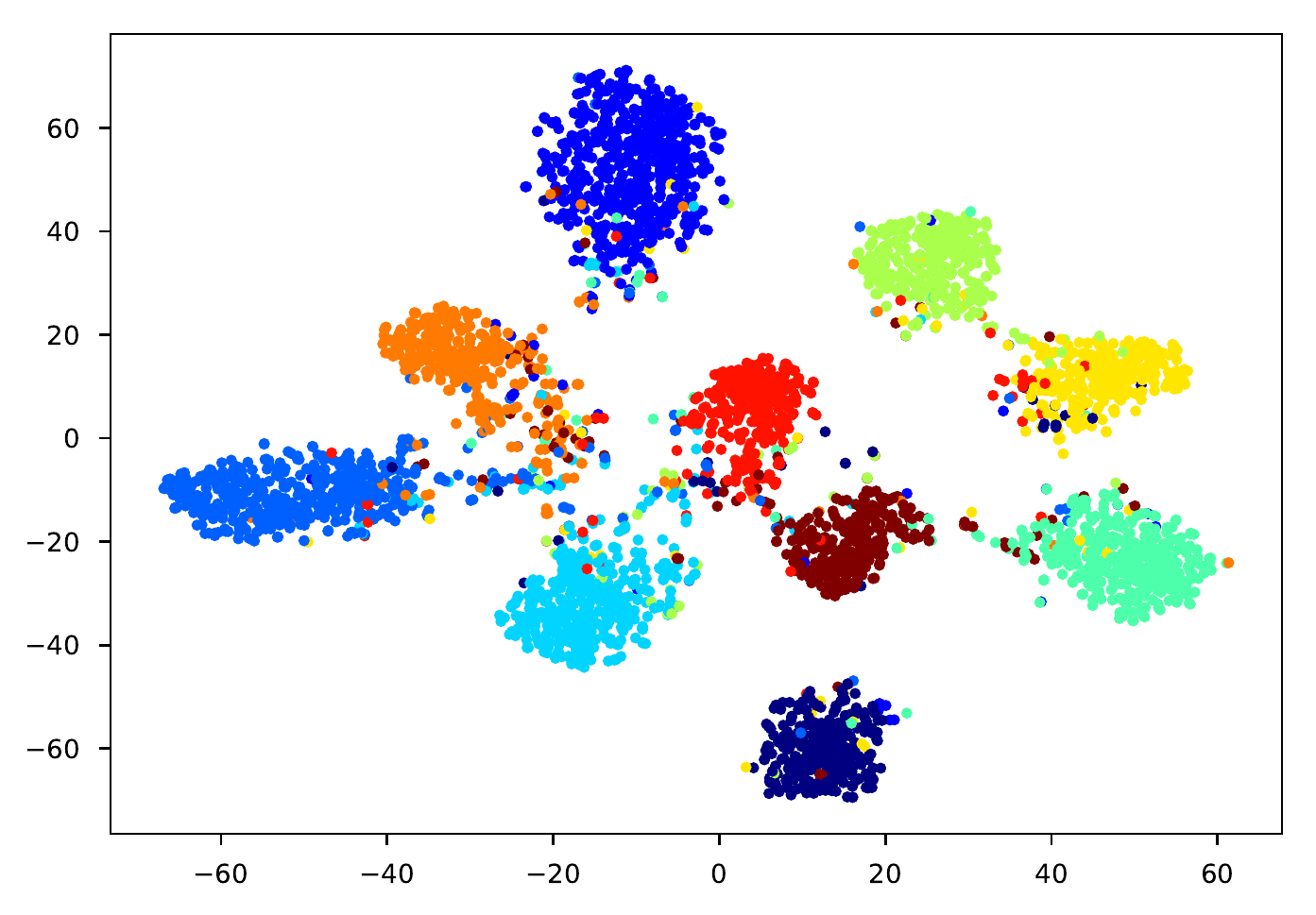}
    \caption{$\mathcal{L}_s$ + $\mathcal{L}_c$ (t-SNE)}
    \label{fig:1}
  \end{subfigure}
  \begin{subfigure}[b]{0.245\textwidth}
    \includegraphics[width=\textwidth]{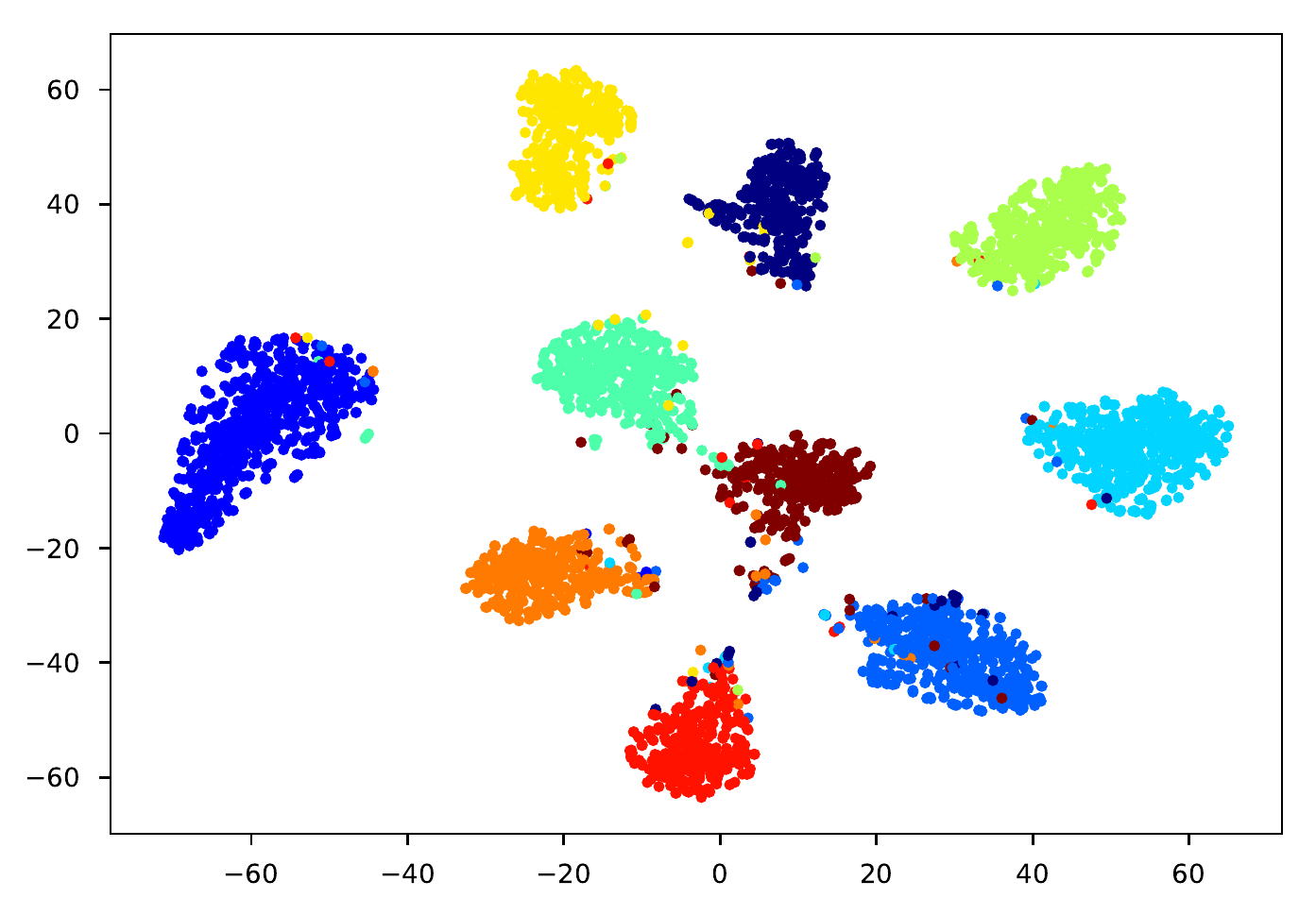}
    \caption{$\mathcal{L}_s$ + $\mathcal{L}_c$ + $\mathcal{L}_d$ (t-SNE)}
    \label{fig:2}
  \end{subfigure}
    \begin{subfigure}[b]{0.248\textwidth}
    \includegraphics[width=\textwidth]{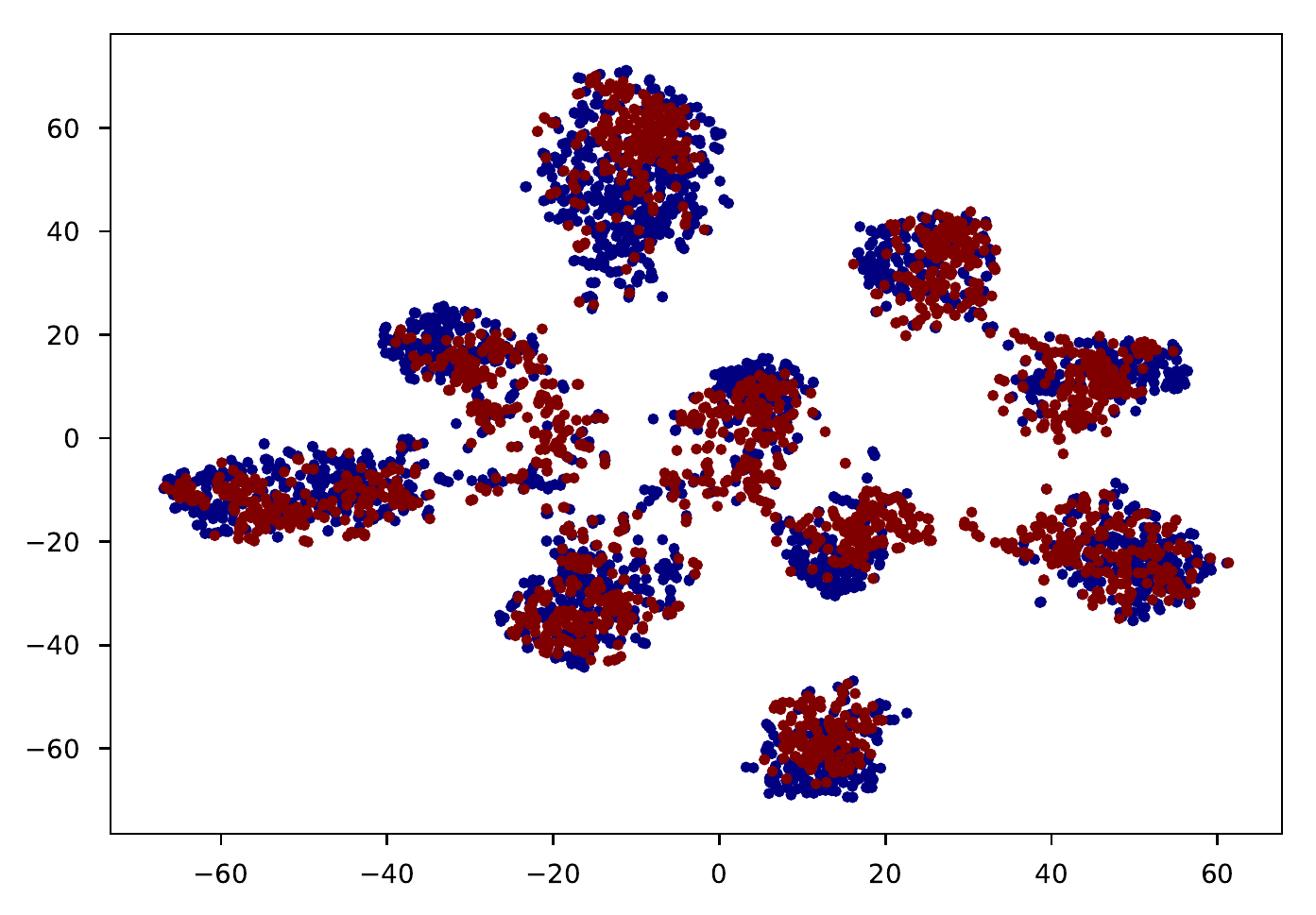}
    \caption{$\mathcal{L}_s$ + $\mathcal{L}_c$ (t-SNE)}
    \label{fig:3}
  \end{subfigure}
  \begin{subfigure}[b]{0.248\textwidth}
    \includegraphics[width=\textwidth]{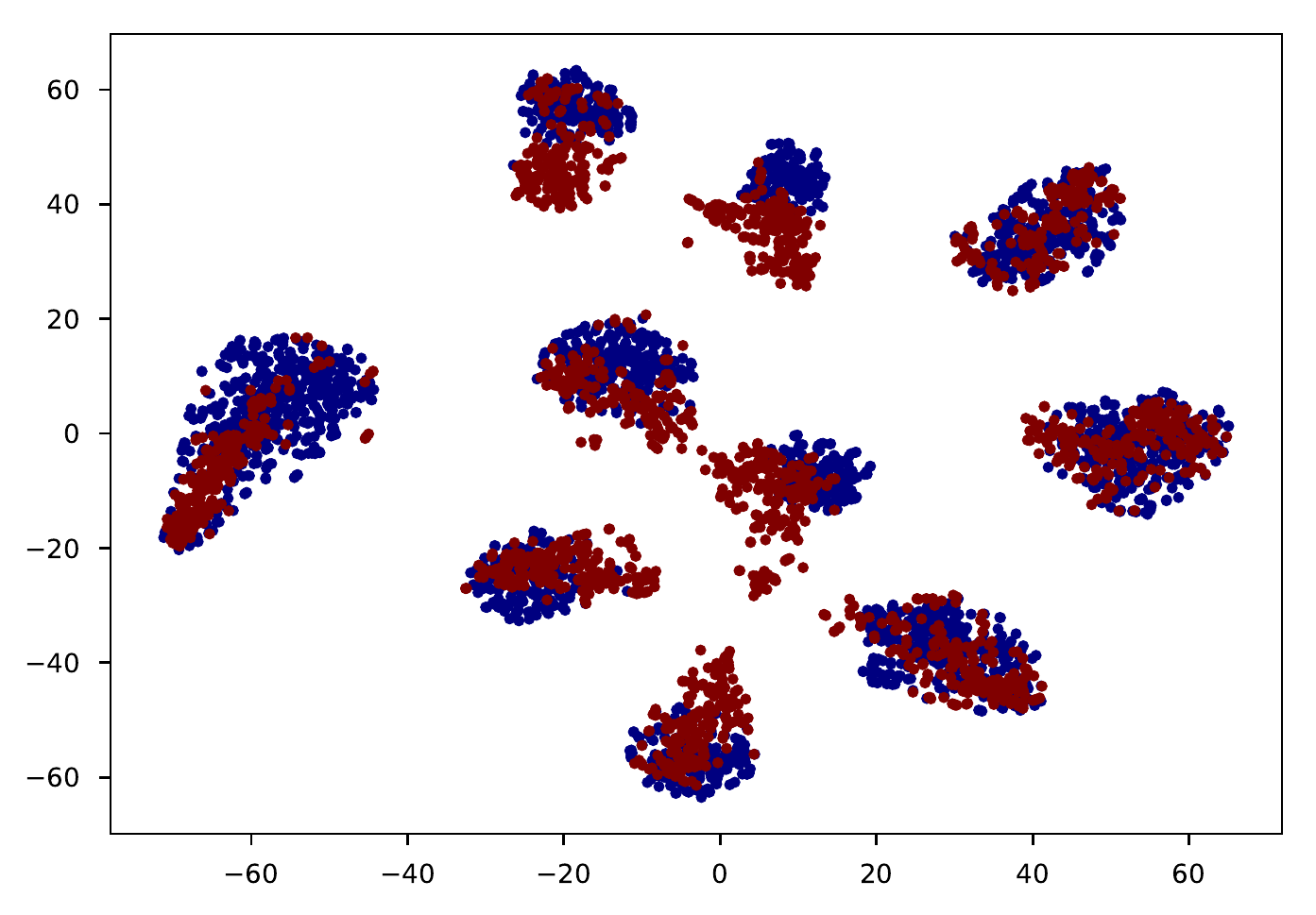}
    \caption{$\mathcal{L}_s$ + $\mathcal{L}_c$ + $\mathcal{L}_d$ (t-SNE)}
    \label{fig:4}
  \end{subfigure}
\caption{The t-SNE visualization of the SVHN$\rightarrow$MNIST task. (a)(b) are generated from category information and each color in (a)(b) represents a category. (c)(d) are generated from domain information. Red and blue points represent samples of source and target domains, respectively.    }
\label{imgTsne}
\end{figure*}

\subsubsection{Feature Visualization}
 To better illustrate the effectiveness of our approach,  we randomly select 2000 samples in the source domain, set the feature (input of the Softmax loss) dimension as 2 and then plot the 2D features in Figure \ref{img2DD}. Compared with the features obtained by methods without the  proposed discriminative loss $\mathcal{L}_d$ (Figure \ref{2D1} and Figure \ref{2D3}), the features obtained by the methods with our discriminative loss (Figure \ref{2D2} and Figure \ref{2D4}) become much more compact and well separated. In particular, the features given by Source Only (Figure \ref{2D1}) are in the form of a strip, while the features given by Source Only with our discriminative loss (Figure \ref{2D2}) are tightly clustered, and there exists a great gap between the clusters. This demonstrates that our proposal can make the model learn more distinguishing features.

The visualizations of the learned features in Figure \ref{img2DD} show great discrimination in the source domain. But this does not mean that our method is equally effective on the target domain. Therefore, we visualize the t-SNE embeddings \cite{donahue2014decaf} of the last hidden layer learned by CORAL or JDDA on transfer task SVHN$\rightarrow$MNIST in Figures \ref{fig:1}-\ref{fig:2} (with category information) and Figures \ref{fig:3}-\ref{fig:4} (with domain information). We can make intuitive observations. \textbf{(1)} Figure \ref{fig:1} ($\mathcal{L}_s$ + $\mathcal{L}_c$) has more scattered points distributed on the inter-class gap than Figure \ref{fig:2} ($\mathcal{L}_s$ + $\mathcal{L}_c$ + $\mathcal{L}_d$) , which suggests that the features learned by JDDA are discriminated much better than that learned by CORAL (larger class-to-class distances). (2) As shown in Figures \ref{fig:3} ($\mathcal{L}_s$ + $\mathcal{L}_c$) and Figures \ref{fig:4} ($\mathcal{L}_s$ + $\mathcal{L}_c$ + $\mathcal{L}_d$), with CORAL features, categories are not aligned very well between domains, while with features learned by JDDA, the categories are aligned much better. All of the above observations can demonstrate the advantages of our approach: whether in the source or target domain, the model can learn more transferable and more distinguishing features with the incorporation of our proposed discriminative loss.
\begin{figure}[!ht]
   \centering
     \includegraphics*[width=1\linewidth]{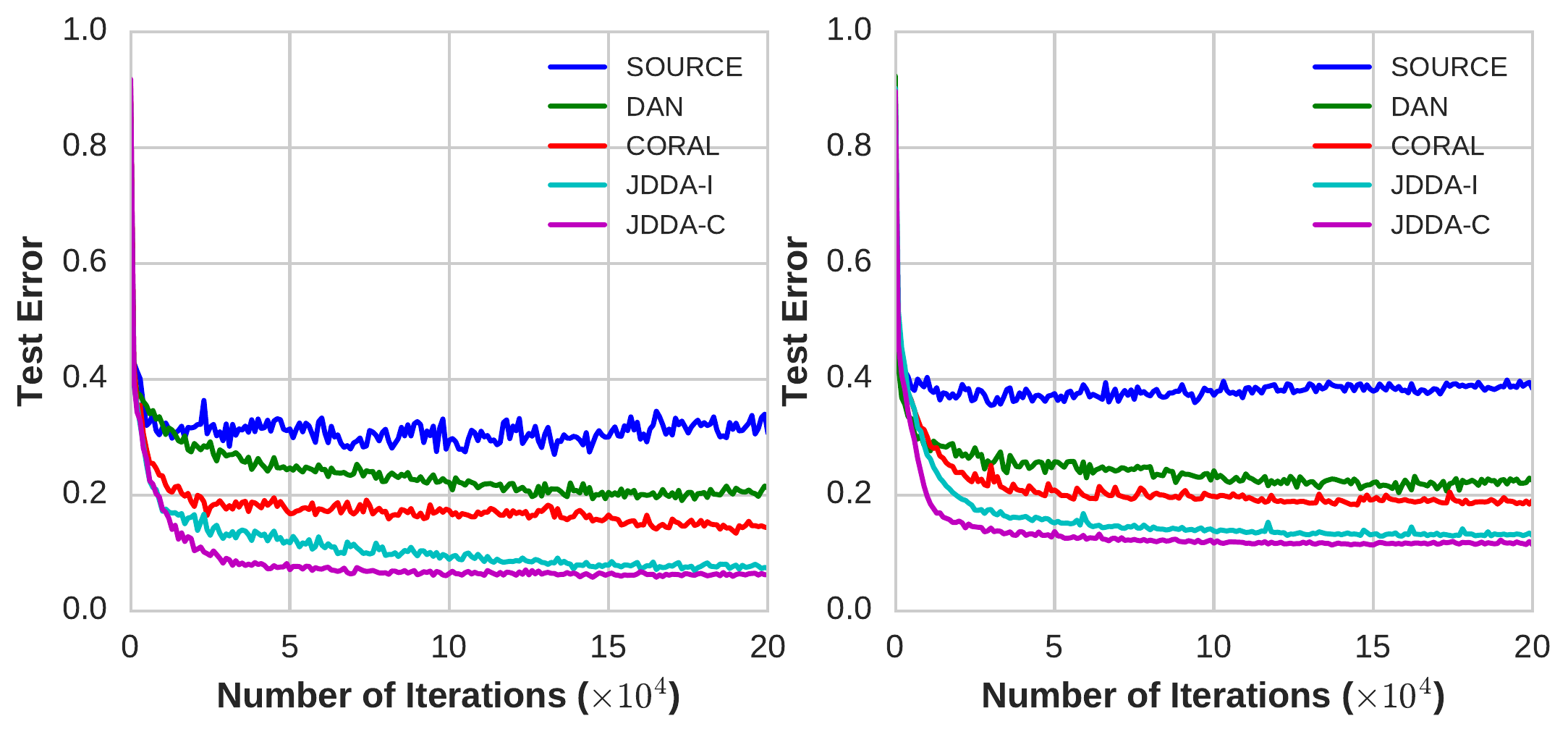}
   \caption{Comparison between JDDA and other start-of-the-art methods in the convergence performance on SVHN$\rightarrow$MNIST (right) and MNIST$\rightarrow$MNIST-M (left).  }
   \label{convergency}
\end{figure}

\subsubsection{Convergence Performance}
 We evaluate the convergence performance of our method through the test error of the training process.  Figure \ref{convergency} shows the test errors of different methods on SVHN$\rightarrow$MNIST and MNIST$\rightarrow$MNIST-M, which reveals that incorporating the proposed discriminative loss helps achieve much better performance on the target domain. What's more, the trend of convergence curve  suggests that JDDA-C converges fastest due to it considers the global cluster information of the domain invariant features during training. In general, our approach converges fast and stably to a lowest test error, meaning it can be trained efficiently and stably to enable better domain transfer.

\subsubsection{Parameter Sensitivity}
\begin{figure}[!ht]
   \centering
     \includegraphics*[width=1\linewidth]{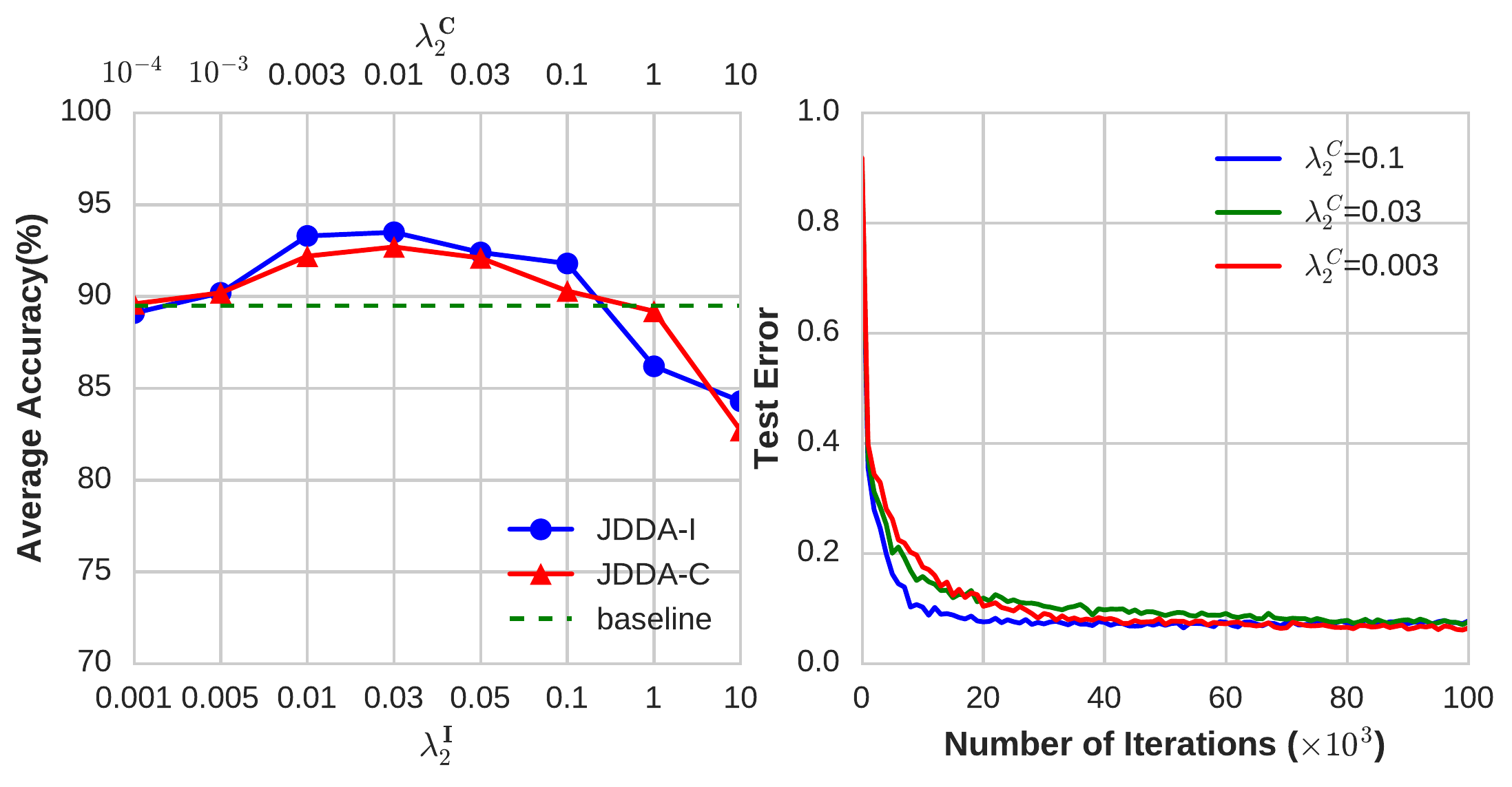}
   \caption{Parameter sensitivity analysis of our approach. The figure on the left shows the Average Accuracy w.r.t. $\lambda_2$ (the $\lambda_2^I$ is the hyper-parameter of JDDA-I and the $\lambda_2^C$ is the hyper-parameter of JDDA-C)  and the figure on the right shows the convergence performance w.r.t. $\lambda_2^C$.  }
   \label{sensi2}
\end{figure}

 We investigate the effects of the parameter $\lambda_2$ which balances the contributions of our proposed discriminative loss. The larger $\lambda_2$ would lead to more discriminating deep features, and vice versa. The left one in the Figure \ref{sensi2} shows the variation of average accuracy as $\lambda_2^I\in{\{0.0001,0.001,0.003,0.01,0.03,0.1,1,10\}}$ or $\lambda_2^C\in{\{0.001,0.005,0.01,0.03,0.05,0.1,1,10\}}$ on task SVHN$\rightarrow$MNIST. We find that the average accuracy increases first and then decreases as $\lambda_2$ increases and shows a bell-shaped curve, which demonstrates a proper trade-off between domain alignment and discriminative feature learning  can improve transfer performance. The right one in the Figure \ref{sensi2} gives an illustration of the relationship between convergence performance and $\lambda_2^C$. We can observe that the model can achieve better convergence performance as $\lambda_2^C$ is appropriately increased. This confirms our motivation that when the speed of feature alignment can keep up with the changing speed of the source domain feature under the influence of our discriminative loss, we can get a domain adaptation model with fast convergence and high accuracy.

\section{Conclusion}
In this paper, we propose to boost the transfer performance by jointing domain alignment and discriminative feature learning. Two discriminative feature learning methods are proposed to enforce the shared feature space with better intra-class compactness and inter-class separability, which can benefit both domain alignment and final classification. There are two reasons that the discriminative-ness of deep features can contribute to domain adaptation. On the one hand, since the shared deep features are better clustered, the domain alignment can be performed much easier. On the other hand, due to the better inter-class separability, there is a large margin between the hyperplane and each cluster. Therefore, the samples distributed near the edge, or far from the center of each cluster in the target domain are less likely to be misclassified. Future researches may focus on how to further mitigate the domain shift in the aligned feature space by other constrains for the domain invariant features.

\bibliographystyle{aaai}
\fontsize{9.5pt}{10.5pt} \selectfont
\bibliography{reference}

\end{document}